\documentclass[letterpaper, 10 pt, conference]{ieeeconf} 

\usepackage{hyperref}
\usepackage{url}
\usepackage{graphics} 
\usepackage{times} 
\usepackage{booktabs}       
\usepackage{amsfonts}       
\usepackage{nicefrac}       
\usepackage{microtype}      
\usepackage{algorithm,multirow,xcolor}
\usepackage{algorithmicx}
\usepackage{algpseudocode}
\usepackage{mathtools}
\usepackage{graphicx}
\usepackage{cases}
\usepackage{array}
\usepackage{color}
\usepackage{float}
\usepackage{amssymb}
\usepackage{subcaption}
\usepackage{wrapfig}
\usepackage{amsmath}
\usepackage{tabularx}
\usepackage{soul}
\usepackage{amsmath}
\usepackage{graphicx} 
\usepackage{epstopdf}
\usepackage{graphicx}
\usepackage{booktabs} 
\usepackage{multirow} 
\usepackage{colortbl} 
\usepackage{paralist} 
\usepackage{setspace}

\IEEEoverridecommandlockouts 
\title{\LARGE \bf
4D-ROLLS: 4D Radar Occupancy Learning via LiDAR Supervision}

\author{Ruihan Liu$^{1\dagger}$,  Xiaoyi Wu$^{1\dagger}$, Xijun Chen$^{1}$,  Liang  Hu$^{1*}$ and Yunjiang Lou$^{1}$ 
\thanks{$\dagger$ Equal Contribution, $*$ Corresponding Author.}
\thanks{$^{1}$R. Liu, X. Wu, X. Chen, L.~Hu and Y. ~Lou are with the Department
of Automation, School of Intelligence Science and Engineering, Harbin Institute of Technology, Shenzhen,
China. For correspondence: ~l.hu@hit.edu.cn.}}
\begin{document}

\maketitle
\begin{abstract}
A comprehensive understanding of 3D scenes is essential for autonomous vehicles (AVs), and among various perception tasks, occupancy estimation plays a central role by providing a general representation of drivable and occupied space. However, most existing occupancy estimation methods rely on LiDAR or cameras, which perform poorly in degraded environments such as smoke, rain, snow, and fog. In this paper, we propose 4D-ROLLS, the first weakly supervised occupancy estimation method for 4D radar using the LiDAR point cloud as the supervisory signal. Specifically, we introduce a method for generating pseudo-LiDAR labels, including occupancy queries and LiDAR height maps, as multi-stage supervision to train the 4D radar occupancy estimation model. Then the model is aligned with the occupancy
map produced by LiDAR, fine-tuning its accuracy in occupancy estimation. Extensive comparative experiments validate the exceptional performance of 4D-ROLLS. Its robustness in degraded environments and effectiveness in cross-dataset training are qualitatively demonstrated. The model is also seamlessly transferred to downstream tasks BEV segmentation and point cloud occupancy prediction, highlighting its potential for broader applications. The lightweight network enables 4D-ROLLS model to achieve fast inference speeds at about 30 Hz on a 4060 GPU. The code of 4D-ROLLS will be made available at https://github.com/CLASS-Lab/4D-ROLLS.
\end{abstract}
\section{introduction}
\begin{figure}[t]
    \centering
    \includegraphics[width=\linewidth]{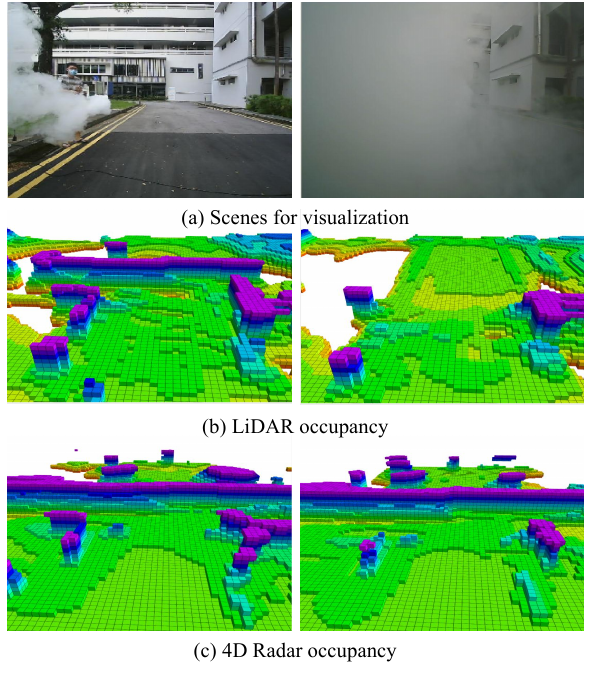}
    \caption{We compare our method with the classic LiDAR-based occupancy estimation approach, ALSO \cite{boulch2023also}, in foggy scenes. Under normal weather conditions, both methods demonstrate strong performance. However, in degraded environments, such as fog, our method remains robust, whereas ALSO fails due to the significant reduction or even complete absence. } 
    \label{fig1}
\end{figure}

Understanding the spatial structure of the world is fundamental for autonomous driving and robotics \cite{fisher2021colmap} \cite{huang2023tri}, where accurate occupancy mapping is crucial for tasks such as environment perception and navigation. In particular, the need for reliable, real-time 3D representations of the environment is especially pressing for robots operating in dynamic, unstructured environments. 
Most existing occupancy estimation methods rely on LiDAR or cameras, which offer precise geometric information and semantic information under ideal conditions \cite{yang2025metaocc}. However, LiDAR and cameras suffer perception degradation easily under adverse weather conditions, as shown in Fig.\ref{fig1}. In comparison, 4D radar that operates at longer wavelengths is resistant to adverse weather conditions such as rain, snow, and fog \cite{efear-4d}, and hence has emerged as an all-weather alternative sensor \cite{get2025}. 

The use of 4D radar in occupancy estimation remains an open challenge. Unlike LiDAR which captures dense and accurate point cloud, 4D Radar point cloud are much sparser and suffer from noises induced by multi-path refection and artifacts \cite{zhuoins20234drvo, zeller2022gaussian}, leading to ambiguous spatial structures.  Additionally, Radar signals can penetrate certain objects, making it challenging to accurately delineate occupied and free space by traditional geometric priors used in LiDAR-based methods, such as ray-tracing \cite{khurana2023point}. Consequently, there is still a research gap in effective 4D radar-based occupancy learning that fully exploits the advantages of 4D radar's sensing capabilities while overcoming its sensing limitations.

To bridge the gap, we propose 4D-ROLLS, a model that uses LiDAR's rich geometric information as supervisory signals to learn 4D radar occupancy representations. 4D-ROLLS achieve comparable performance to LiDAR occupancy estimation while being lightweight and scalable.
 By combining a sparse 3D point cloud encoder with a 2D backbone, and utilizing LiDAR projection constraints, we achieve a weakly-supervised radar occupancy estimation model that can be fine-tuned for various robotic tasks without requiring large amounts of labeled data. This approach demonstrates high-quality occupancy estimation with 4D radar, approaching the precision of LiDAR, while maintaining the lightweight and generalizable nature required for real-world robot applications. The contributions of the research are threefold:
\begin{itemize}
        \item We propose 4D-ROLLS, a novel method for 4D radar-based occupancy field estimation that leverages the resilience of 4D radar to various weather conditions, while effectively addressing the challenges brought by sparsity and inherent noise in radar data;
    \item We introduce a novel method for constructing pseudo-labels from LiDAR point cloud across space. This approach enhances the accuracy and reliability of the labels and avoids the high costs associated with manual annotation. To the best of our knowledge, this is the first method to develop a weakly supervised occupancy model for 4D radar, relying solely on LiDAR as the supervisory signal;
    \item Our model achieves robust performance across various scenarios, providing a universal and high-quality environment representation for downstream robotic tasks like BEV segmentation and 3D occupancy prediction.
\end{itemize}
\section{related works}
\subsection{4D radar perception}
In recent years, there has been a growing effort to explore the potential of 4D radar in autonomous driving applications, including 3D object detection \cite{palffy2022multi} \cite{wang2022interfusion}, 3D reconstruction \cite{huang2024dart}, odometry \cite{efear-4d} \cite{zhuang20234d} and occupancy prediction \cite{ding2024radarocc}. Although traditional geometric methods such as Bayesian occupancy filter \cite{ronecker2024dynamic} have been applied to 4D radar, they are usually based on manual rules and restricted by the noise of 4D radar, leading to limited application. With recent advancements in deep learning, there has been a shift towards data-driven approaches in the broader perception and environmental understanding tasks, such as occupancy estimation. Many successful methods have been primarily developed for LiDAR-based \cite{zuo2023pointocc} and vision-based \cite{wei2023surroundocc} \cite{tan2024geocc} techniques, leveraging dense and high-quality sensor data to model and predict occupancy in structured environments.

Despite significant progress in related fields, 4D radar occupancy estimation remains relatively underdeveloped. Some recent efforts have sought to bridge this gap. Ming et al. \cite{ming2024occfusion} and Ma \cite{ma2024licrocc} proposed Occfusion and LiCRocc, respectively, combining 2D radar with cameras and LiDAR to achieve precise 3D occupancy predictions. Both methods utilize a shared backbone for LiDAR and 2D radar. However, due to the sparsity of 4D radar point cloud, the LiDAR backbone cannot be directly applied to 4D radar-based systems. Ding et al. \cite{ding2024radarocc} innovatively introduced Radarocc, the first 4D radar-based 3D occupancy prediction method. While Radarocc demonstrates promising results, it can only apply to input in raw 4D radar tensors rather than in radar point cloud. Currently, many 4D radar dataset provide only 4D radar point clouds but not lower-level frequency signals in tensors, such as the ones provided in the public datasets VOD \cite{apalffy2022}, NTU \cite{zhang2023ntu4dradlm} and MSC \cite{choi2023msc}.   Such observation motivates us to develop 3D occupancy estimation model with 4D radar point cloud as input.
\begin{figure*}[t]
    \centering
    \includegraphics[width=\textwidth]{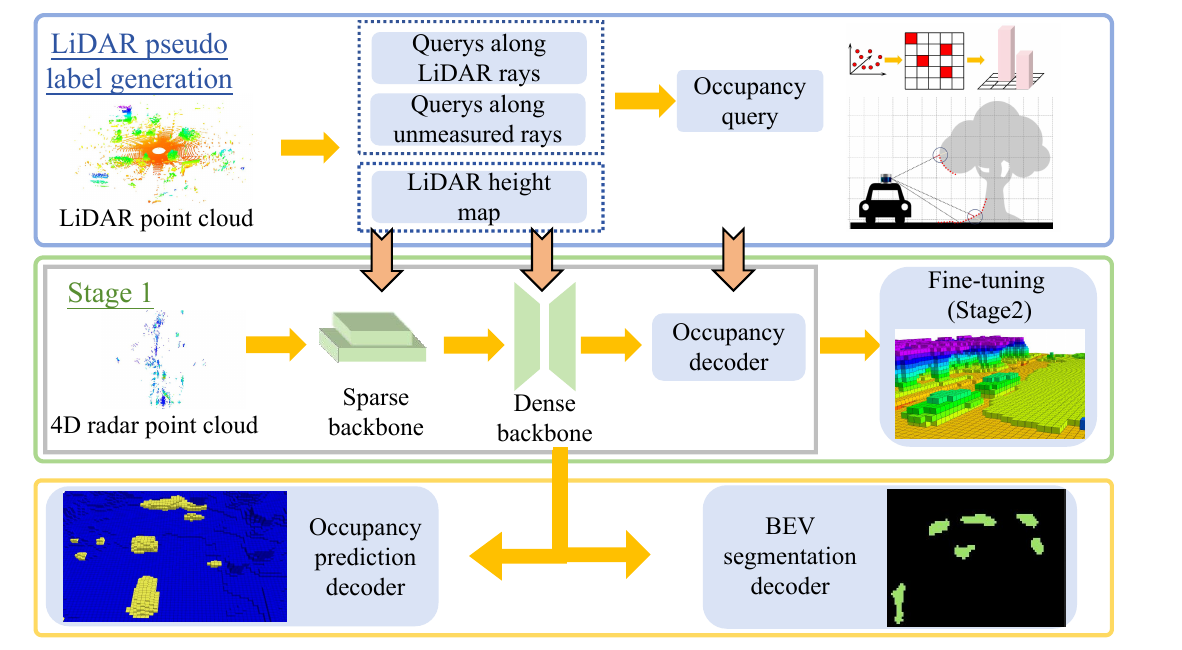}
    \caption{Overview of the Framework: including Occupancy Estimation, Prediction Results, and BEV Segmentation Performance}
    \label{Framework}
\end{figure*}
\subsection{LiDAR-based occupancy estimation}
3D occupancy prediction advances scene understanding through dense geometric-semantic representations. UDNet \cite{zou2021up} employs 3D U-Nets but faces computational overhead due to empty voxels. L-CONet \cite{wang2023openoccupancy} introduces a coarse-to-fine supervision strategy to alleviate the heavy computational burden. Inspired by TPVFormer \cite{huang2023tri}, PointOcc \cite{zuo2023pointocc} transforms LiDAR point cloud into a cylindrical TPV representation, which is then used to effectively restore the 3D structured representation and obtain high-resolution 3D occupancy predictions. While these existing methods demonstrate excellent accuracy, they are often limited in practical applications due to the high cost and difficulty of acquiring labeled data, especially for sensors like 4D radar, which lack large-scale labeled datasets. To overcome these challenges, self-supervised learning has emerged as a promising alternative. Approaches like \cite{boulch2023also} and \cite{agro2024uno} eliminate the need for manual annotations by leveraging the inherent structure of the data, enabling models to learn fine-grained, complete representations from large-scale, unlabeled datasets. 

Inspired by these self-supervised methods, we propose to leverage the rich, fine-grained structure of LiDAR data as weak supervision to train the 4D radar occupancy model. By learning from LiDAR’s dense and well-structured representations, even sparse radar data can be used effectively to infer high-quality scene understanding. 
\section{methodology}
A system overview is depicted in Fig. \ref{Framework}, which presents a weakly supervised learning approach for 4D radar occupancy that leverages LiDAR point cloud as the supervisory signal. In Section III.A, we introduce our LiDAR pseudo label generation method for supervising. In Section III.B, we detail the occupancy network structures. Finally, in Section III.C, we showcase fine-tuning for various downstream tasks.
\subsection{LiDAR pseudo label generation}
\label{label_generation}
Our pseudo label generation includes two parts: the first one is occupancy query generation, and the second one is LiDAR height map generation. 

\textbf{Occupancy query generation: }Building on Alexandre's method \cite{boulch2023also}, we conduct our occupacy query generation method. Given a LiDAR point cloud, let \textit{P} denote the set of points captured by the LiDAR scan. Since LiDAR operates with a short wavelength, its points lie on the surfaces of objects. Therefore, we can assume that all voxels along the line from the LiDAR sensor to each point are unoccupied. Based on this assumption, we generate two types of queries: occupied and unoccupied points.

Let \textit{c} be the location of LiDAR and $p_i$ be the $i_{th}$ point of point cloud. The unit direction vector from the LiDAR center to $p_i$ is defined as: $u=\frac{p_i-c}{||p_i-c||}$. We define two query points, $R^{+}$ and $R^{-}$, where $R^{+}$ represent the occupied label and $R^{-}$ represent the unoccupied label. For a single point 
$p_i$, the corresponding sampling points $R^{+}_{i}$ and $R^{-}_{i}$ can be expressed as:
\begin{align*}
R^{+}_{i}=p_i+ru, r>0 \\
R^{-}_{i}=p_i-ru, r>0 
\end{align*}
Here, $R^{+}_{i}$ is considered as an occupied point, located in a small range $r$ behind $p_{i}$ as long as the object observed is not thinner than $r$, while $R^{-}_{i}$ is regarded as an unoccupied point, lying in the opposite direction considered as empty from our assumption. And the combination of these two queries constitutes our occupancy query as a supervision.

\textbf{LiDAR height map generation:} Although 4D radar point cloud contains height information, the generated point cloud typically exhibits a discrete, clustered distribution along the Z-axis due to the relatively low spatial resolution and significant measurement noise. This results in a lack of well-defined geometric structures, hindering the accurate representation of object shapes and details. Therefore, it is necessary to use the high geometric accuracy of LIDAR to maximize the feature accuracy of 4D radar in the Z-axis during training. Inspired by a well-known method in SLAM \cite{kim2018scan}, we propose our LiDAR height map as an assisted supervision of our occupancy estimation. 

Specifically, we first construct a height map from the LiDAR point cloud, where the space is divided into xy-plane voxel grids, and the maximum Z-axis value within each voxel is recorded as a feature, similar to the concept of Scan Context \cite{kim2018scan}. During training, we project the radar point cloud into the LiDAR coordinate system and identify the XY-plane voxels containing radar points. For these voxels, we retrieve the corresponding maximum Z-value from the LiDAR height map as a pseudo label to supervise the network output. This supervision mechanism effectively transfers LiDAR geometric priors to radar occupancy estimation, thereby improving the model’s ability to represent the environment against the sparse and noisy nature of 4D radar point cloud.

\subsection{Occupancy network}
The occupancy network mainly consists of the sparse encoder, dense encoder-decoder and occupancy head.

\textbf{Sparse encoder: }For 4D radar point cloud, their inherent sparsity poses challenges for feature extraction. When using 3D voxelization-based networks such as \cite{zhou2018voxelnet}, a significant proportion of voxels remain empty, leading to inefficient spatial feature extraction. Additionally, due to the extensive detection range of 4D radar (up to 300m), voxelization demands a large computational space, resulting in high memory and processing costs.
To address these issues, we adopt the backbone from \cite{zuo2023pointocc}, which projects 4D radar point cloud onto three 2D perpendicular planes and extracts point-wise features. Specifically, it employs a point-wise MLP for feature encoding, voxelizes the points with MaxPooling, and utilizes spatial group pooling to extract three-plane features—\( F_{BEV} \), \( F_{FV} \), and \( F_{SV} \). These representations capture complementary spatial information: \( F_{BEV} \) encodes horizontal structure by compressing features along the Z-axis, while \( F_{FV} \) and \( F_{SV} \) retain vertical and lateral details, facilitating robust occupancy estimation. 

Although the TPV representation effectively captures multi-view spatial features, its reliance on discrete 2D projections may lead to the loss of fine-grained depth information. To address this, we introduce LiDAR height map (illustrated in Section.\ref{label_generation}) as a constraint to enhance geometric accuracy along the Z-axis. Because of the low vertical resolution of 4D radar, the front-view (\( F_{FV} \)) and side-view (\( F_{SV} \)) projections lack sufficient detail and are prone to errors. Therefore, we utilize the BEV feature map (\( F_{BEV} \)) to estimate the height map. 
Specifically, a \(1 \times 1\) convolutional layer is employed to aggregate information across the channel dimension. Given an input feature tensor \( F_{BEV} \in \mathbb{R}^{B \times C \times H \times W} \), where \( B \) denotes the batch size, \( C \) represents the number of feature channels, and \( H \) and \( W \) correspond to the spatial dimensions, the predicted height map \( H_{\text{pred}} \in \mathbb{R}^{B \times 1 \times H \times W} \) is computed as  
\begin{equation}
    H_{\text{pred}} = \sigma(W * F_{BEV} + b)
\end{equation}
where \( W \in \mathbb{R}^{1 \times C \times 1 \times 1} \) and \( b \in \mathbb{R} \) are the learnable weights and bias, respectively, \( * \) denotes the convolution operation, and \( \sigma(\cdot) \) represents ReLU activation function.  
And we employ a Mean Squared Error (MSE) loss between the predicted height map \( H_{\text{pred}} \) and the LiDAR-guided height map \( H_{\text{LiDAR}} \). Given that the radar point cloud is inherently sparse, we compute the MSE loss only at valid spatial locations where radar provides observations.
Formally, let \( M \in \{0,1\}^{H \times W} \) be a binary mask indicating valid locations where radar points are present, with \( M_{h,w} = 1 \) if a radar return exists at spatial position \( (h, w) \), and \( M_{h,w} = 0 \) otherwise. The weighted MSE loss is then defined as:

\begin{equation}
    \mathcal{L}_{\text{height}}^1 = \frac{1}{\sum_{h,w} M_{h,w}} \sum_{h,w} M_{h,w} \cdot (H_{\text{pred}, h,w} - H_{\text{LiDAR}, h,w})^2.
\end{equation}

This formulation ensures that the supervision signal is applied only to regions where radar measurements are available, preventing misleading gradients from unobserved areas. By leveraging LiDAR as a supervisory signal, the model learns to refine its height predictions, leading to a more accurate occupancy representation. Notably, although the height map is calculated, it will not be used as input to the subsequent networks.

\textbf{Dense Decoder:} After encoding the 4D radar point cloud into a TPV representation, we utilize a TPV-based backbone, where the encoder, adapted from an image backbone, extracts multi-scale features for each plane. The decoder aggregates these features to reconstruct high-resolution TPV representations, namely \(T_{BEV}\), \(T_{FS}\), and \(T_{SV}\). Inspired by the deep supervision technique used in BEV segmentation \cite{li2024bevformer}, we refine the height estimation by applying a \(1 \times 1\) convolution on the aggregated features. This introduces a deeper supervision signal as an additional loss term, denoted as $\mathcal{L}_{\text{height}}^2$.

\textbf{Occupancy Head:} The final occupancy head is designed by aggregating features from three views. After sampling, these features are weighted and fused into a unified feature representation. This unified feature map is then passed through a decoder to estimate the occupancy map, generated by LiDAR. The final loss is computed between the predicted occupancy map and the occupancy queries, denoted as $\mathcal{L}_{\text{occ}}$. The overall loss for stage 1 is defined as:
\begin{equation}
    \mathcal{L} = \omega_1 * \mathcal{L}_{\text{height}}^1 + \omega_2 * \mathcal{L}_{\text{height}}^2 + \omega_3 * \mathcal{L}_{\text{occ}}
\end{equation}

\textbf{Fine-tuning: }To enhance the inference ability of 4D radar occupancy estimation, we propose the second stage training process in which a self-supervised LiDAR model is used to fine-tune a 4D radar-based occupancy network. First, a self-supervised model is trained solely using LiDAR point cloud, where we leverage the spatial structure and richness of LiDAR data to learn a robust occupancy representation. In the second stage, the radar model is fine-tuned using the LiDAR-augmented supervision with the same network structure. This fine-tuning process allows the 4D radar model to correct its initial predictions by aligning them with the occupancy map produced by LiDAR, thereby removing spurious or false occupancy estimates, such as walls or objects that the radar sensor fails to detect. By leveraging this LiDAR self-supervision, the model can generalize better to 4D radar data, improving its robustness in environments where radar coverage is incomplete or noisy. 
\subsection{Downstream tasks}
In this section, to show the representation capacity and transferability of 4D-ROLLS, we transfer it to effectively perform the downstream tasks of BEV segmentation and point cloud occupancy prediction via simple fine-tuning methods.

For each pixel in the BEV plane, we first aggregate the occupancy features corresponding to that pixel across all layers of the feature map. A MLP is employed to compute a weight for each occupancy feature, and the weighted features are then summed to generate a consolidated BEV feature representation. This procedure can be viewed as a simplified attention-based pooling mechanism, where the most relevant features for each BEV pixel are weighted and combined. The resulting feature map, now enriched with spatial context, is then fed into a U-Net \cite{ronneberger2015u} architecture for a specific head, both BEV segmentation and occupancy prediction.
\section{experiments}
\begin{figure*}[t]
    \centering
     \includegraphics[width=\textwidth]{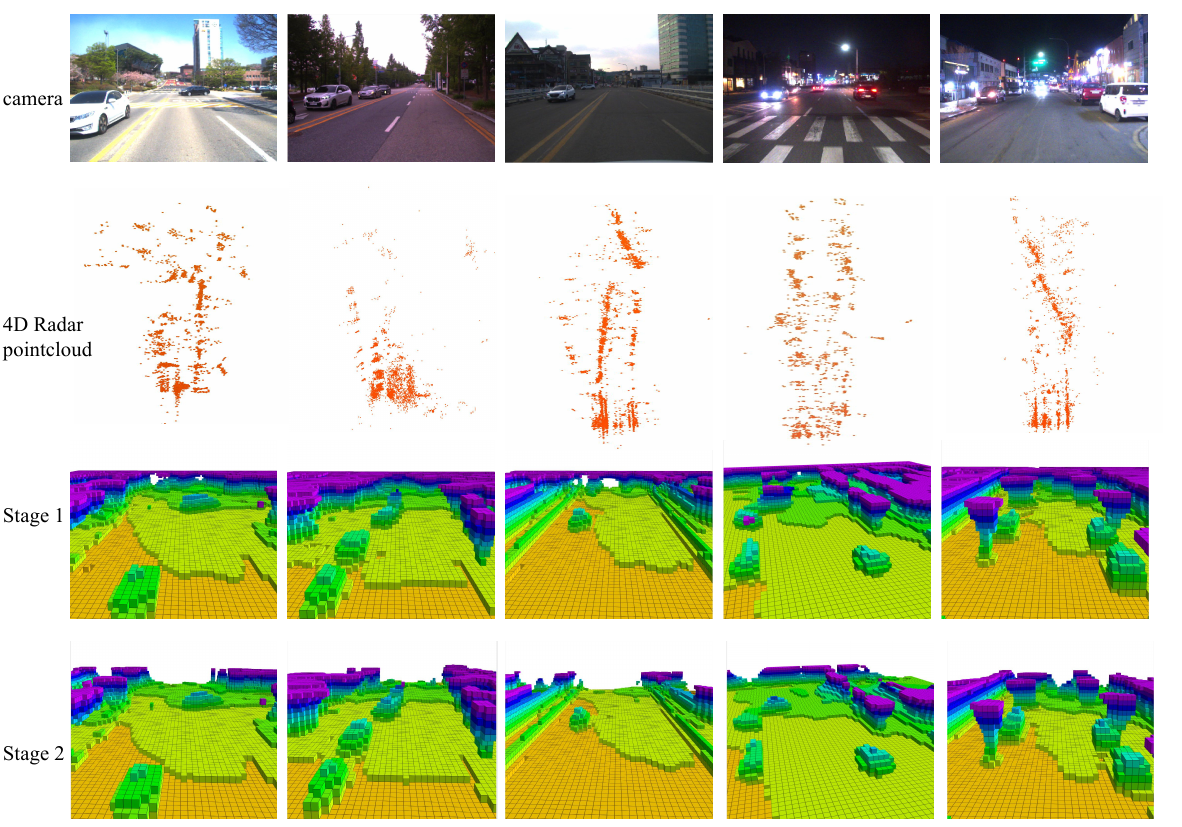}
    \caption{Qualitative comparison on the MSC dataset. Despite the sparse 4D radar point cloud, our approach effectively infers fine-grained scene details. Our method achieves promising results in the Stage 1. Through fine-tuning, it further refines the predictions, correcting points that the 4D radar previously failed to capture.}
    \label{msc_pic}
\end{figure*}
\begin{table*}[t]
\centering
\caption{QUANTITATIVE COMPARISON ON MSC DATASET}
\renewcommand\arraystretch{1.25}{
\begin{tabularx}{\textwidth}{p{1.8cm}XXXXXXXXXXXXXXXXXX}
\hline
Sequences& \multicolumn{3}{c}{URBAN\_{}F0} & \multicolumn{3}{c}{LOOP\_{}C0} & \multicolumn{3}{c}{URBAN\_{}B0} & \multicolumn{3}{c}{RURAL\_{}B0} & \multicolumn{3}{c}{RURAL\_{}B1} \\
\cline{1-16}
& CD $\downarrow$ & AR $\downarrow$ & L2 $\downarrow$ & CD $\downarrow$ & AR $\downarrow$ & L2 $\downarrow$ & CD $\downarrow$ & AR $\downarrow$ & L2 $\downarrow$ & CD $\downarrow$ & AR $\downarrow$ & L2 $\downarrow$ & CD $\downarrow$ & AR $\downarrow$ & L2 $\downarrow$\\
\hline
ALSO-L$^{*}$&1.252&0.150&3.690&1.497&0.165&3.901&0.576&0.128&2.057&0.983&0.146&3.590&1.071&0.150&3.265\\
UNO-L&1.405&0.171&4.136&1.278&0.174&4.182&0.663&0.149&2.408&0.893&0.158&3.849&1.146&0.165&3.578\\
\hline
ALSO-R$^{**}$&3.030&0.195&\textbf{4.662}&1.958&0.165&3.828&0.915&0.144&2.209&1.707&0.182&3.879&1.752&0.178&3.490\\
UNO-R&5.932&0.228&5.735&4.826&0.264&5.902&1.347&0.174&2.967&2.260&0.198&4.778&2.480&0.200&4.490\\
Ours-stage1&2.983&\textbf{0.193}&4.664&1.858&0.174&4.074&\textbf{0.762}&0.137&2.129&1.479&\textbf{0.166}&3.788&\textbf{1.380}&\textbf{0.165}&3.468\\
Ours-stage2&\textbf{2.792}&0.198&4.692&\textbf{1.547}&\textbf{0.161}&\textbf{3.769}&0.787&\textbf{0.136}&\textbf{2.117}&\textbf{1.423}&0.168&\textbf{3.754}&1.411&0.167&\textbf{3.461}\\

\hline
\end{tabularx}
{\footnotesize 
{*} xxx-L and {**} xxx-R  refer to the original LiDAR occupancy method with LiDAR as input and that adapted with 4D radar as input, repectively.
}
}
\label{msc}
\end{table*}
\subsection{Experimental setup}
\textbf{Dataset: }Since 4D radar remains an emerging technology, currently there are no publicly available datasets as comprehensive or well-established as the KITTI dataset \cite{geiger2012we}, which offers diverse all-weather scenes and detailed annotations. Although the VoD dataset \cite{apalffy2022} provides labeled annotations, its 4D radar scans are too sparse, limiting its suitability to object detection or sensor fusion rather than 4D radar-solely occupancy estimation. Similarly, the K-Radar dataset \cite{paek2022k}, collected from raw radar signals, suffers from significant noise in its tensor data and large memory requirements.
Hence, we utilize the MSC \cite{choi2023msc} and NTU \cite{zhang2023ntu4dradlm} datasets for training and testing. The MSC dataset comprises 90,000 frames, with several thousand points per frame on average, and the NTU dataset serves as a supplementary source. Despite the rich scene diversity, both datasets lack annotated labels. Therefore, we first evaluated occupancy estimation accuracy through reconstruction and subsequently generated our own labels for downstream tasks using an image segmentation algorithm. 

\textbf{Training details: }
The estimation range is limited to (0, 51.2) m, (-25.6, 25.6) m and (-3, 3) m along the X, Y and Z axis, respectively. The occupancy voxel size is (0.4m, 0.4m, 0.4m). For dense backbone, we leverage UNO-decoder \cite{agro2024uno} to keep simple and efficiency. We employ the AdamW optimizer with a weight decay of 0.01 and an initial learning rate of 4e-4 for stage 1, 1e-4 for fine-tuning. All experiments are conducted on a NVIDIA 3090 GPU with a batch size 16 for 7 epochs for stage 1 and batch size 2 for fine-tuning. And followed by \cite{agro2024uno}, we utilize Chamfer distance (CD), Near Field Chamfer Distance (NFCD), L2 error and relative L2 error (AR) as metrics to measure our geometric reconstruction accuracy. 

\begin{figure*}
    \includegraphics[width=\textwidth]{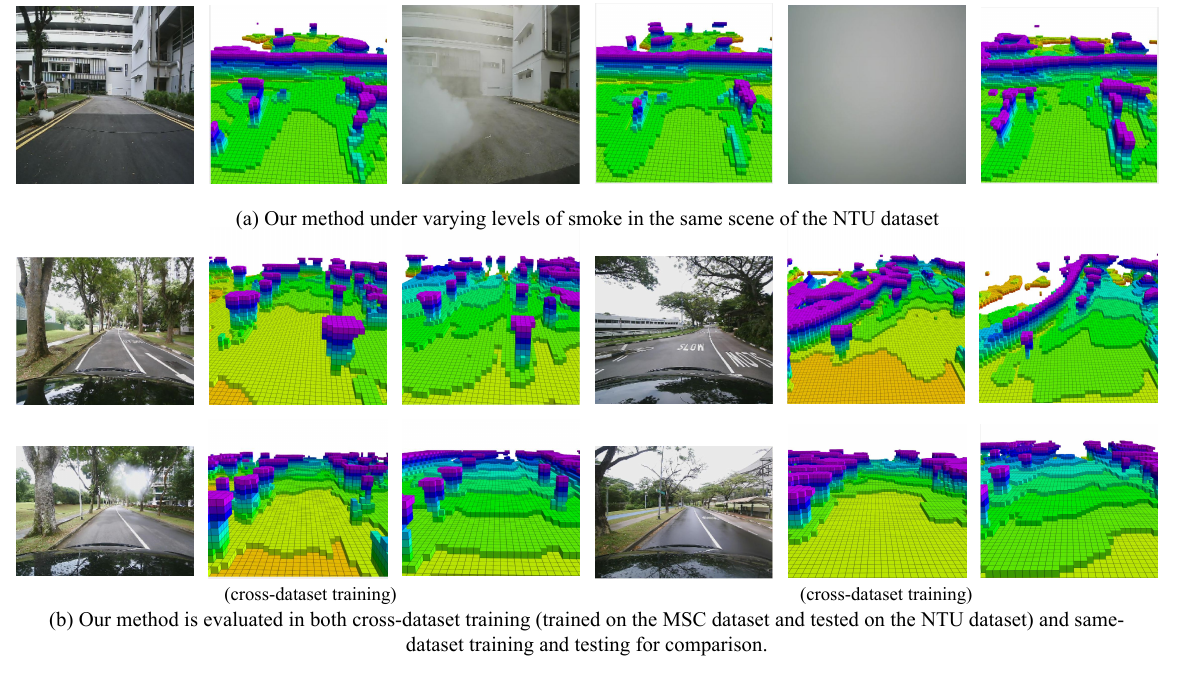}
    \caption{Qualitative comparison on NTU dataset, including cross-dataset testing and intra-dataset testing.}
    \label{ntu_pic}
\end{figure*}
\begin{table*}[t]
\centering
\caption{QUANTITATIVE COMPARISON ON NTU DATASET}
\renewcommand\arraystretch{1.25}{
\begin{tabularx}{\textwidth}{p{1.8cm}XXXXXXXXXXXXX}
\hline
Sequences& \multicolumn{4}{c}{cp} & \multicolumn{4}{c}{loop1} & \multicolumn{4}{c}{loop2}  \\
\hline
& CD $\downarrow$ &NFCD $\downarrow$& AR $\downarrow$ & L2 $\downarrow$ & CD $\downarrow$ &NFCD $\downarrow$& AR $\downarrow$ & L2 $\downarrow$ & CD $\downarrow$ &NFCD $\downarrow$& AR $\downarrow$ & L2 $\downarrow$\\
\hline
ALSO-L$^{*}$&0.686&0.522&0.281&3.411&0.731&0.467&0.177&2.693&0.765&0.512&0.196&2.757\\
UNO-L&0.548&0.606&0.320&4.172&0.800&0.612&0.215&3.379&1.304&0.810&0.237&3.564\\
\hline
ALSO-R$^{**}$&1.542&1.277&0.311&3.759&1.387&1.101&\textbf{0.201}&3.308&1.685&1.435&0.247&3.504&\\
UNO-R&1.363&\textbf{1.068}&0.371&4.749&1.642&1.280&0.237&3.863&2.166&1.439&0.260&3.879\\
Ours-stage1&1.198&1.085&0.298&3.606&0.970&0.794&0.209&3.277&1.280&\textbf{1.033}&\textbf{0.223}&3.142\\
Ours-stage2&\textbf{1.036}&1.106&\textbf{0.298}&\textbf{3.542}&\textbf{0.846}&\textbf{0.777}&0.202&\textbf{3.120}&\textbf{1.103}&1.232&0.227&\textbf{3.117}\\

\hline
\end{tabularx}
{\footnotesize 
{*} xxx-L and {**} xxx-R  refer to the original LiDAR occupancy method with LiDAR as input and that adapted with 4D radar as input, repectively.
}
}
\label{ntu}
\end{table*}
\subsection{Comparison against state-of-the-art on datasets}
Tab.\ref{msc} shows the comparison results of our methods and baselines on MSC dataset. Because there is currently no open-source 4D radar point cloud solely based occupancy algorithms, we modified two SOTA LiDAR occupancy algorithms, Uno \cite{agro2024uno} and ALSO \cite{boulch2023also} to accommodate 4D radar input as baseline for comparisons. The original LiDAR algorithms are used as references. Our 4D radar-based occupancy estimation significantly outperforms the baseline methods derived from LiDAR occupancy in all key metrics, demonstrating the effectiveness of our LiDAR-supervised radar occupancy learning strategy. Despite the significant improvements, a performance gap remains when compared to LiDAR-based methods in normal weather. Radar signals are affected by multipath reflections and penetrability, and its point cloud resolution is significantly lower than that of LiDAR, limiting its ability to capture fine-grained environmental details. While our approach incorporates LiDAR-supervised height estimation, the absence of explicit geometric priors in radar data can still lead to misclassification of free space and object boundaries in complex scenes, ultimately affecting the quality of occupancy prediction.  

The results of ours 4D-ROLLS with only stage 1 (Our-stage1) and that with stage 2 as well (Our-stage2) demonstrate the impact of fine-tuning with LiDAR supervision. Although the quantitative metrics show only marginal improvement, Fig. \ref{msc_pic} reveals that falsely predicted occupied regions in Stage 1 caused by the absence of 4D radar detections are effectively pruned in Stage 2. This effect is particularly noticeable in areas where the radar point cloud is sparse such as the far side, leading to a more accurate and physically plausible occupancy representation. This underscores the advantage of incorporating LiDAR supervision, which mitigates the limitations of 4D radar-based perception while preserving its all-weather capability. However, in extreme cases, radar may capture objects after penetration, while the LiDAR supervision signal completely ignores them, causing the network to learn a biased world model.

Tab. \ref{ntu} presents the comparison results on the NTU dataset. Although the number of frames in this dataset is limited, its abundance of structured environments enhances 4D radar perception, making it more suitable for our approach. As a result, our method shows a significant performance improvement over the baseline, exceeding ALSO and UNO by more than 30\%{}. Additionally, though this dataset uses Livox LiDAR for supervision, which is considerably sparser compared to the 128-line Ouster LiDAR used in the MSC dataset, our method still maintains high accuracy.

Fig. \ref{ntu_pic} (a) illustrates the performance of our method under varying levels of smoke in nearly identical scenes from the NTU dataset. Since LiDAR degrades in smokes and becomes unreliable as ground truth, we can only qualitatively demonstrate the stability of our method. Our occupancy model remains reliable and robust in foggy environments, underscoring its potential for adverse weather conditions.

Furthermore, to evaluate the generalization capability of our approach, we conducted a cross-dataset test, training the model on the MSC dataset and testing it on the NTU dataset, as shown in Fig. \ref{ntu_pic}(b). The results indicate that our method performs comparably well in the cross-dataset test, achieving accuracy close to that obtained when training directly on the NTU dataset. Notably, we observed a consistent tilt in the predictions when training exclusively on the NTU dataset, which may be attributed to a slight misalignment in the NTU radar’s physical installation.

\begin{figure}
    \centering
    \includegraphics[width=\linewidth]{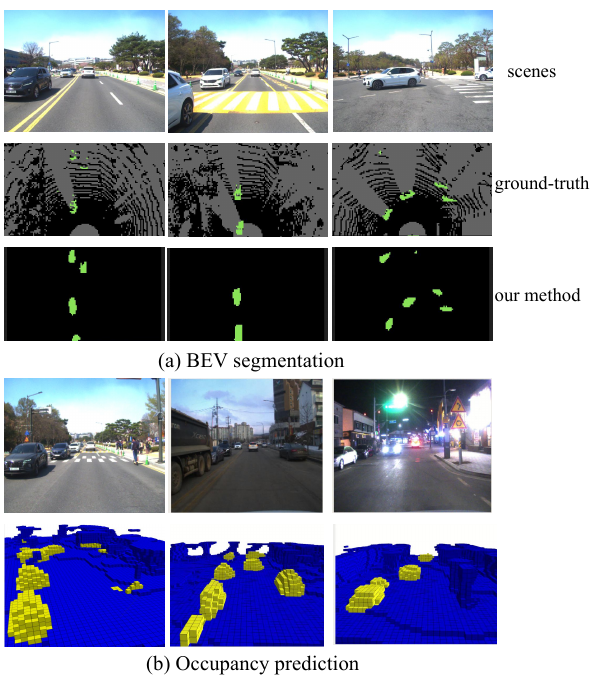}
    \caption{The results of downstream tasks using our method.}
    \label{downstream_pic}
\end{figure}
\subsection{Downstream task results}
In this subsection, we demonstrate the effectiveness of pre-training for occupancy in both 3D occupancy prediction and BEV segmentation, as illustrated in Fig.\ref{downstream_pic}. Since the MSC dataset lacks annotated labels, we use HRNet \cite{sun2019deep} to generate pseudo-labels from images for fine-tuning the 3D occupancy prediction. Additionally, we employ Point-labelers from the Semantic KITTI dataset \cite{behley2019semantickitti} to supervise the BEV segmentation task. Our method successfully detects all vehicles in the scene, as well as some pedestrians, despite only very few points being reflected from them by 4D radar. This is likely due to the mapping from radar to LiDAR being established during pre-training, which enables our method to achieve good results even when a simple approach is used for downstream tasks.  
\subsection{System efficiency}
We evaluate the inference speed of our proposed method on a Jeston orin equipped with an NVIDIA 4060 GPU. The average inference time per step for occupancy estimation is 33.08 ms, that is at the frequency of 30 Hz. This performance shows our method's potential for real-time application in robot navigation and other tasks.
\begin{table}[t]
    \centering
    \caption{System efficiency of different task for one step inference}
    \begin{tabular}{cc}
    \hline
        Tasks & Time (ms) \\
        \hline
         Occupancy estimation& 33.08 \\
         Occupancy prediction&33.30 \\
         BEV segmentation&43.62\\
         \hline
    \end{tabular}
    \label{time_tab}
\end{table}
\section{CONCLUSION}
In this paper, we propose 4D-ROLLS, a novel and lightweight method for 4D radar-only occupancy estimation with weak supervision from LiDAR. First, we introduce a technique to generate pseudo-LiDAR labels, including occupancy queries and LiDAR height maps. Next, we employ a sparse backbone and a dense backbone to generate occupancy estimates. To further enhance the robustness of the model in complex environments, we fine-tune the model using an unsupervised LiDAR-based occupancy model, which helps eliminate spurious or false occupancy estimations. Finally, we leverage our occupancy model as a pre-trained model to efficiently perform downstream tasks, including BEV segmentation and occupancy prediction, demonstrating the versatility of our approach. Extensive experiments demonstrate the effectiveness, real-time performance, and strong generalization of our method across various tasks.  Future work will explore the extension of its capabilities to more complex scenarios and real-world applications. 
\bibliographystyle{IEEEtran}
\bibliography{ref}

\end{document}